\documentclass[11pt]{article}
\usepackage{acl2015}
\usepackage{times}
\usepackage{url}
\usepackage{latexsym}
\usepackage{graphicx}
\usepackage{multirow}
\usepackage{amssymb,amsfonts}
\title{Multichannel Variable-Size Convolution for Sentence Classification}
\author{Wenpeng Yin \rm{and} \textbf{Hinrich Sch\"{u}tze}\\
	    Center for Information and Language Processing\\University of Munich, Germany\\
	    {\tt wenpeng@cis.uni-muenchen.de}}\author{Wenpeng Yin \rm{and} \textbf{Hinrich Sch\"{u}tze}\\
	    Center for Information and Language Processing\\University of Munich, Germany\\
	    {\tt wenpeng@cis.uni-muenchen.de}}
\date{}

\def\tabref#1{Table~\ref{tab:#1}}

\def\eqref#1{Eq.~\ref{eqn:#1}}

\newcounter{notecounter}
\newcommand{\enotesoff}{\long\gdef\enote##1##2{}}
\newcommand{\enoteson}{\long\gdef\enote##1##2{{
\stepcounter{notecounter}
\large\bf
\hspace{1cm}\arabic{notecounter} $<<<$ ##1: ##2
$>>>$\hspace{1cm}}}}
\enoteson
\enotesoff

\begin{document}
\maketitle

\begin{abstract}
We propose \emph{MVCNN}, a convolution neural network
(CNN) architecture for sentence classification.  
It (i) combines diverse versions of pretrained
word embeddings and (ii) extracts features of multigranular
phrases with variable-size convolution
filters.  We also show that \emph{pretraining}
MVCNN is critical for good performance.  MVCNN achieves
state-of-the-art performance on four tasks: on small-scale
binary, small-scale multi-class and large-scale Twitter
sentiment prediction and on subjectivity classification.
\end{abstract}

%This paper comes up with a new model \emph{MVCNN} based on
%convolution neural network (CNN) in order to promote the
%accuracy of typical sentence classification tasks. Despite
%intensive utilization of CNN in Natural Language Processing,
%the main contributions of MVCNN include: a) combining
%\textbf{M}ultiple pretrained embedding versions; b)
%extracting variable-size phrase features by
%\textbf{V}ariable-size convolution filters. In addition,
%this work uses two training tricks to enhance the
%performance of MVCNN: \emph{mutual-learning} of embedding
%versions and \emph{pretraining} of MVCNN.  We test the MVCNN
%in four experiments: small scale binary and multi-class
%sentiment prediction, large-scale Twitter sentiment
%prediction and subjectivity classification. The network gets
%state-of-the-art performance in all tasks.

\section{Introduction}\label{sec:intro}
Different sentence classification tasks are crucial for many
Natural Language Processing (NLP) applications. Natural
language sentences have complicated structures, both
sequential and hierarchical, that are essential for
understanding them. In addition, how to decode and
compose the features of component units,
including single words and variable-size phrases, is central
to the sentence classification problem.

In recent years, deep learning models have achieved
remarkable results in computer vision
\cite{krizhevsky2012imagenet}, speech recognition
\cite{graves2013speech} and NLP
\cite{collobert2008unified}. 
A problem largely specific to NLP is how to detect features of linguistic units, how to
conduct composition over variable-size sequences and how to use them for
NLP tasks
\cite{collobert2011natural,blunsom2014convolutional,kimEMNLP2014}. 
\newcite{socher2011dynamic} proposed recursive neural
networks  to form phrases based on parsing trees.
This approach depends on
the availability of a well performing parser; for many
languages and domains, especially noisy domains, reliable parsing is difficult.
Hence, convolution
neural networks (CNN) are getting increasing attention, for they
are able to model long-range dependencies in sentences via
hierarchical structures
\cite{dos2014deep,kimEMNLP2014,denil2014modelling}. Current
CNN systems usually implement a convolution layer with
fixed-size filters (i.e., feature detectors), in which the concrete filter size is a
hyperparameter. They essentially split a sentence into
multiple sub-sentences by a sliding window, then determine
the sentence label by using the dominant label across all
sub-sentences. The underlying assumption is that the sub-sentence
with that granularity is potentially good enough to represent the whole
sentence. However, it is hard to 
find the granularity of a 
``good sub-sentence'' that works well across sentences. This motivates us to
implement \emph{variable-size filters} in a convolution layer
in order to extract features of \emph{multigranular
  phrases}.

Breakthroughs of deep learning in NLP are also based on
learning distributed word representations -- also
called ``word embeddings'' -- by neural language models
\cite{bengio2003neural,mnih2009scalable,mikolov2010recurrent,mikolov2012statistical,mikolov2013efficient}.
Word
embeddings are derived by projecting words from a sparse,
1-of-$V$ encoding ($V$: vocabulary size) onto a
lower dimensional and dense vector space via hidden layers
and can be interpreted as feature extractors that encode semantic and
syntactic features of words.

Many papers study the comparative performance of different
versions of word embeddings, usually learned by different neural
network (NN) architectures.  For example,
\newcite{chen2013expressive} compared HLBL
\cite{mnih2009scalable}, SENNA \cite{collobert2008unified},
Turian \cite{turian2010word} and Huang
\cite{huang2012improving}, showing great variance in quality
and characteristics of the semantics captured by the tested
embedding versions. \newcite{hill2014not} showed that embeddings
learned by neural machine translation models outperform
three representative monolingual embedding versions: skip-gram
\cite{mikolov2013distributed}, GloVe
\cite{pennington2014glove} and C\&W
\cite{collobert2011natural} in some cases.  These prior studies
motivate us to explore combining multiple versions of word
embeddings, treating each of them as a distinct
description of words. Our expectation is that the
combination of these embedding versions, trained by different
NNs on different corpora, should contain more information
than each version individually. We want to leverage this
diversity of different embedding versions to extract higher
quality sentence features and thereby improve sentence classification
performance.

The letters ``M'' and ``V'' in the name
``MVCNN'' of our architecture denote the multichannel and 
variable-size convolution filters, respectively.  ``Multichannel'' employs language from computer
vision where a color image has red, green and blue
channels. Here, a channel is a description by an embedding version.

For many sentence classification tasks, only relatively
small training sets are available.
MVCNN has a large number of parameters, so that overfitting
is a danger when they are trained on small training sets.
We address this problem 
by \emph{pretraining}  MVCNN on unlabeled data.
These pretrained
weights can then be fine-tuned for the specific
classification task.

In sum, we attribute the success of MVCNN to: 
(i) designing variable-size
convolution filters to extract variable-range features of
sentences and
(ii) exploring
the combination of multiple public embedding versions to
initialize words in sentences. We also employ two ``tricks''
to further enhance system performance: mutual learning and pretraining.

In remaining parts, Section \ref{sec:relatedwork} presents
related work. Section \ref{sec:des} gives details of
our classification model. Section \ref{sec:prom} introduces
two tricks that enhance system performance: mutual-learning and pretraining. Section
\ref{sec:exp} reports experimental results. Section
\ref{sec:conc} concludes this work.
\section{Related Work}\label{sec:relatedwork}
Much prior work has
exploited deep neural networks to model sentences. 

\newcite{blacoe2012comparison} represented a sentence by
element-wise addition, multiplication, or recursive
autoencoder over embeddings of component single
words. \newcite{yin2014exploration} extended this
approach by composing on words and phrases
instead of only single words.

\newcite{collobert2008unified} and \newcite{yu2014deep} used
one layer of convolution over phrases detected by a sliding
window on a target sentence, then used max- or
average-pooling to form a sentence representation.

\newcite{blunsom2014convolutional} stacked multiple layers
of \emph{one-dimensional} convolution by dynamic k-max pooling to model
sentences. We also adopt dynamic k-max pooling while our convolution layer has variable-size filters.

\newcite{kimEMNLP2014} also studied multichannel
representation and variable-size filters. Differently, their
multichannel relies on a single version of pretrained
embeddings (i.e., pretrained Word2Vec embeddings) with two
copies: one is kept stable and the other one is fine-tuned by
backpropagation. We develop this insight by incorporating
diverse embedding versions. Additionally, their idea
of variable-size filters is further developed.

\newcite{le2014distributed} initialized the representation
of a sentence as a
parameter vector, treating it as a global feature and
combining this vector with the representations of context
words to do word prediction. Finally, this fine-tuned vector
is used as representation of this sentence. Apparently, this
method can only produce generic sentence representations
which encode no task-specific features.

Our work is also inspired by studies that compared
the performance of different word embedding versions or
investigated the combination of them. For example,
\newcite{turian2010word} compared Brown clusters, C\&W
embeddings and HLBL embeddings in NER and chunking tasks. They
found that Brown clusters and word embeddings both can
improve the accuracy of supervised NLP systems; and
demonstrated empirically that \emph{combining different word
representations is beneficial}. \newcite{luo2014pre} adapted
CBOW \cite{mikolov2013efficient} to train word embeddings on
different datasets: free text documents from Wikipedia,
search click-through data and user query data, showing that
\emph{combining them gets stronger results than using individual
word embeddings} in web search ranking and word similarity
task. However, these two papers either learned word
representations on the same corpus \cite{turian2010word} or
enhanced the embedding quality by extending training
corpora, not learning algorithms \cite{luo2014pre}. In our
work, there is no limit to the type of embedding versions we can
use and
they leverage not only the diversity of corpora, but also
the different principles of learning algorithms.

\section{Model Description}\label{sec:des}

\begin{figure}[t]
\centering
\includegraphics[height=6.5cm,angle=90]{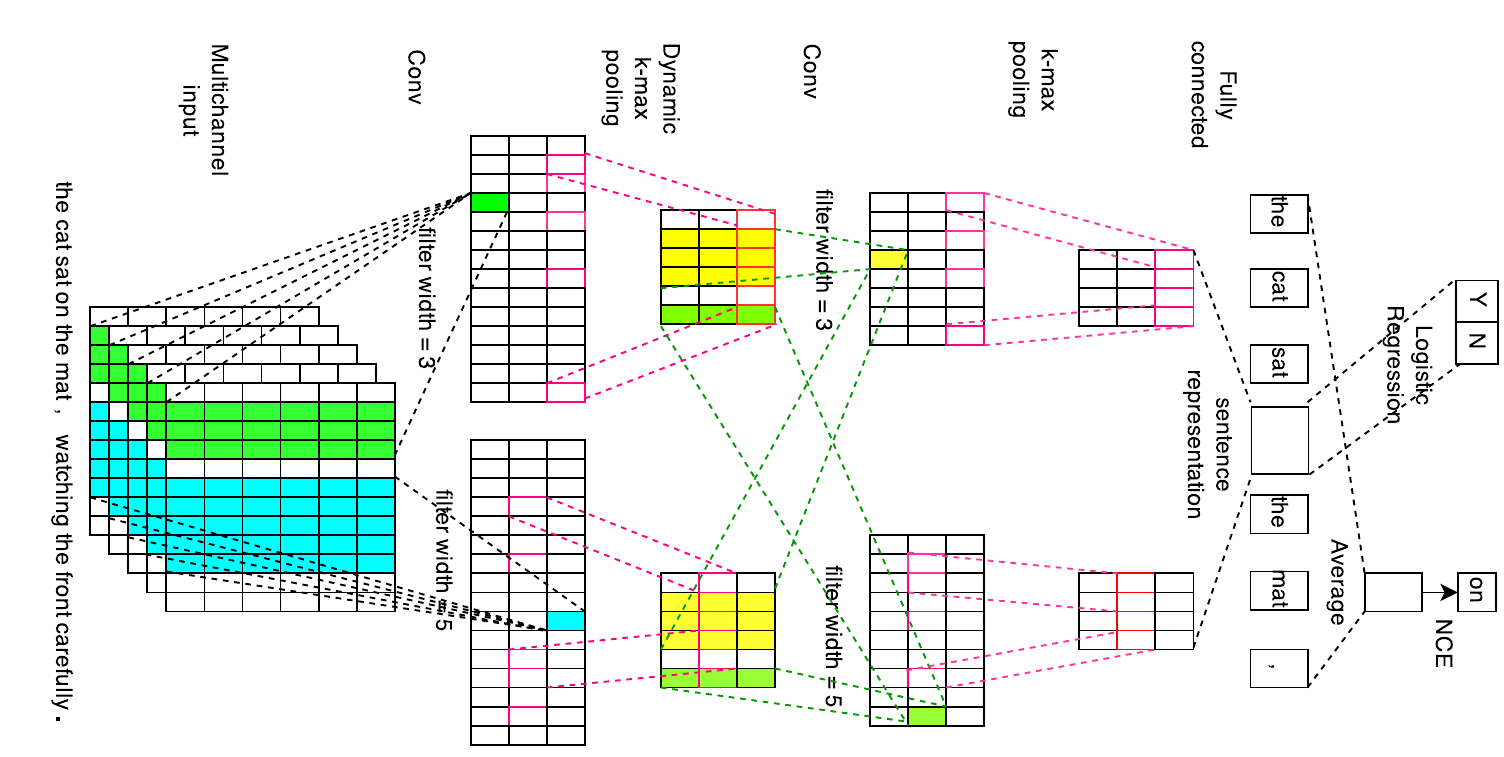}
\caption{MVCNN: supervised classification and pretraining.}
\label{fig:cnn}
\end{figure}

We now describe the architecture of our model
\emph{MVCNN}, illustrated in Figure \ref{fig:cnn}.

\textbf{Multichannel Input.} The input of MVCNN
includes \emph{multichannel feature maps} of a considered sentence, each is a matrix
initialized by a different
embedding version. Let  $s$  be sentence length,
$d$ dimension of word embeddings and $c$ the total number of
different embedding versions (i.e., channels). Hence, the whole initialized input is a
three-dimensional array of size $c\times d\times s$. Figure
\ref{fig:cnn} depicts a sentence with $s=12$ words. Each
word is initialized by $c=5$ embeddings, each coming from a
different channel. In implementation, sentences in a
mini-batch will be padded to the same length, and unknown
words for corresponding channel are randomly initialized or
can acquire good initialization from the
\emph{mutual-learning} phase described in next section.

Multichannel initialization brings two advantages: 1) a
frequent word can have $c$ representations in the beginning
(instead of only one), which means it has more available
information to leverage; 2) a rare word missed in some
embedding versions can be ``made up'' by others (we call it
``partially known word''). Therefore, this kind of
initialization is able to make use of information about
partially known words, without having to employ full random
initialization or removal of unknown words.  The vocabulary
of the binary sentiment prediction task described in
experimental part contains 5232 words unknown in HLBL
embeddings, 4273 in Huang embeddings, 3299 in GloVe
embeddings, 4136 in SENNA embeddings and 2257 in Word2Vec
embeddings. But only 1824 words find no embedding from any
channel! Hence, multichannel initialization can considerably
reduce the number of unknown words.

\textbf{Convolution Layer (Conv).} For convenience, we first
introduce how this work uses a convolution layer on one input
feature map to generate one higher-level feature map. Given
a sentence of length $s$: $w_1, w_2, \ldots, w_s$;
$\mathbf{w}_i\in\mathbb{R}^{d}$ denotes the embedding of
word $w_i$; a convolution layer uses sliding \emph{filters} to
extract local features of that sentence. The filter width
$l$ is a parameter. We first concatenate the initialized
embeddings of $l$ consecutive words ($\mathbf{w}_{i-l+1},
\ldots, \mathbf{w}_i$) as $\mathbf{c}_i\in\mathbb{R}^{ld}$
$(1\leq i <s+l)$, then generate the feature value of this
phrase as $\textbf{p}_i$ (the whole vector
$\textbf{p}\in\mathbb{R}^{s+l-1}$ contains all the
local features) using a tanh activation function and a
linear projection vector $\mathbf{v}\in\mathbb{R}^{ld}$ as:
\begin{equation}
\mathbf{p}_i=\mathrm{tanh}(\mathbf{v}^\mathrm{T}\mathbf{c}_i+b)
\end{equation}

More generally, convolution operation can deal with multiple
input feature maps and can be stacked to yield feature maps
of increasing layers. In each layer, there are usually
multiple filters of the same size, but with different weights
\cite{blunsom2014convolutional}. 
We refer to a filter with a specific set of weights
as a \emph{kernel}. The goal is often to train a model in which
different kernels 
detect different kinds of features of a local region.  However,
this traditional way can not detect the features of regions
of different granularity. Hence we keep the property of
multi-kernel while extending it to variable-size in the same
layer.

As in CNN for object recognition, to
increase the \emph{number} of kernels of a certain
layer, multiple feature maps may be computed in parallel at
the same layer. Further, to increase the \emph{size
  diversity} of kernels in the same layer, more feature maps
containing various-range dependency features can be
learned. We denote a feature map of the $i^{\mathrm{th}}$
layer by $\mathbf{F}_i$, and assume totally $n$ feature maps
exist in layer $i-1$: $\mathbf{F}_{i-1}^1, \ldots,
\mathbf{F}_{i-1}^n$. Considering a specific filter size $l$
in layer $i$, each feature map $\mathbf{F}_{i,l}^j$ is computed by convolving a distinct set of filters of
size $l$, arranged in a matrix $\mathbf{V}_{i,l}^{j,k}$,
with each feature map $\mathbf{F}^k_{i-1}$ and summing the
results:

\begin{equation}
\mathbf{F}_{i,l}^j=\sum^n_{k=1}\mathbf{V}_{i,l}^{j,k}*\mathbf{F}^k_{i-1}
\end{equation}
where $*$ indicates the convolution operation and $j$ is the
index of a feature map in layer $i$. The weights in
$\mathbf{V}$ form a rank 4 tensor.

Note that we use \emph{wide convolution} in this work: it
means word representations $\mathbf{w}_g$ for $g\leq 0$ or
$g\geq s+1$ are actually zero embeddings. Wide convolution
enables that each word can be detected by all filter weights
in $\mathbf{V}$.

In Figure \ref{fig:cnn}, the first convolution layer deals
with an input with $n=5$ feature maps.\footnote{A reviewer
  expresses surprise at such a small number of
  maps. However, we will use four variable sizes (see below), so that
  the overall number of maps is 20.
We use a small number of maps partly because training times
for a network are on the order of days, so limiting the
number of parameters is important.} Its filters have
sizes 3 and 5 respectively (i.e., $l=3, 5$), and each filter
has $j=3$ kernels.
This
means this convolution layer can detect three kinds of
features of phrases with length 3 and 5, respectively.

DCNN in \cite{blunsom2014convolutional} used one-dimensional
convolution: 
each higher-order feature is produced from values of a
single dimension in the lower-layer feature map.
 Even though that work proposed
\emph{folding} operation to model the dependencies between
adjacent dimensions, this type of dependency modeling is still 
limited. Differently, convolution in present work is able to
model dependency across dimensions as well as adjacent
words, which obviates the need for a folding
step. This change also means our model has substantially
fewer parameters than the DCNN since the output of each
convolution layer is smaller by a factor of $d$.

\enote{hs}{was:

DCNN in \cite{blunsom2014convolutional} used one-dimensional
convolution: 
the same dimension in a feature map is mapped
to a high-order feature. Even though that work proposed
``folding'' operation to model the dependencies between
adjacent dimensions, the dependency modeling is still pretty
limited. Differently, convolution in present work is able to
model dependency across dimensions as well as adjacent
words, which obviates the need for their ``folding''
step. This change also means our model has substantially
fewer parameters than the DCNN, since the output of each
convolution layer is smaller by a factor of $d$.

}

\textbf{Dynamic k-max Pooling.} 
\newcite{blunsom2014convolutional} pool the $k$
most active features compared with simple max (1-max) pooling
\cite{collobert2008unified}. This property enables it to
connect multiple convolution layers to form a deep
architecture to extract high-level abstract features. In
this work, we directly use it to extract features for
variable-size feature maps. For a given feature map  in
layer $i$, dynamic k-max pooling
extracts $k_{i}$ top values from each dimension and
$k_{top}$ top values in the top layer. We set
\begin{equation}
\nonumber
k_{i}=\mathrm{max}(k_{top}, \lceil\frac{L-i}{L}s\rceil)
\label{equ:kdy}
\end{equation}
where $i\in\{1,2,\ldots\, L\}$ is the order of convolution
layer from bottom to top in Figure \ref{fig:cnn}; $L$ is the
total numbers of convolution layers; $k_{top}$ is a
constant determined empirically, we set it to 4 as
\cite{blunsom2014convolutional}.

As a result, the second convolution layer in Figure
\ref{fig:cnn} has an input with two same-size feature maps,
one results from filter size 3, one from filter size 5.  The
values in the two feature maps are for phrases with
different granularity. The motivation of this convolution
layer lies in that a feature reflected by a short phrase may
be not trustworthy while the longer phrase containing the
short one is trustworthy, or the long phrase has no
trustworthy feature while its component short phrase is more
reliable.  This and even higher-order convolution layers
therefore can \emph{make a trade-off} between the features of different granularity.

\textbf{Hidden Layer.} On the top of the final k-max pooling, we stack a fully connected layer to learn sentence representation with given dimension (e.g., $d$). 

\textbf{Logistic Regression Layer.} Finally, sentence representation is forwarded into logistic regression layer for classification. 

In brief, our MVCNN model learns from \cite{blunsom2014convolutional} to use dynamic k-max pooling to stack multiple convolution layers, and gets insight from \cite{kimEMNLP2014} to investigate variable-size filters in a convolution layer. Compared to \cite{blunsom2014convolutional}, MVCNN has rich feature maps as input and as output of each convolution layer. Its convolution operation is not only more flexible to extract features of variable-range phrases, but also able to model dependency among all dimensions of representations. MVCNN extends the network in \cite{kimEMNLP2014} by hierarchical convolution architecture and further exploration of multichannel and variable-size feature detectors. 

\section{Model Enhancements}\label{sec:prom}
This part introduces two training tricks that enhance
the performance of MVCNN in practice.

\textbf{Mutual-Learning of Embedding Versions.}
One observation in using multiple embedding versions is that they
have different vocabulary coverage.
An unknown word in an embedding version may be
a known word in another version. Thus, there exists a proportion of words that can only be partially initialized by certain versions of word embeddings, which means these words lack the description from other versions. 

To alleviate
this problem, we design a \emph{mutual-learning} regime to
predict representations of unknown words for each embedding
version by learning projections between versions.
As a result, 
all
embedding versions have the same vocabulary.
This processing ensures that more words in each embedding version receive a good
representation, and is expected to give most words occurring in a classification dataset more comprehensive initialization (as opposed to just being randomly initialized).

Let $c$ be the number of embedding versions in consideration,
$V_1, V_2, \ldots, V_i, \ldots, V_c$ their vocabularies,
$V^*=\cup^c_{i=1} V_i$ their union, and
$V_i^-=V^*\backslash V_i$ ($i=1, \ldots, c$) the vocabulary
of unknown words for embedding version $i$. Our goal is to learn
embeddings for the words in $V_i^-$ by knowledge from
the other $c-1$ embedding versions.

We use the
overlapping vocabulary between $V_i$ and $V_j$,
denoted as $V_{ij}$, as training set, formalizing a projection $f_{ij}$ from space
$V_i$ to space $V_j$ ($i\neq j; i,
j\in\{1,2,\ldots,c\}$) as follows:
\begin{equation}
\mathbf{\hat{w}}_j=\mathbf{M}_{ij}\mathbf{w}_i
\label{equ:proj}
\end{equation}
where
$\mathbf{M}_{ij}\in\mathbb{R}^{d\times d}$,
$\mathbf{w}_i\in\mathbb{R}^d$ denotes the representation of
word $w$ in space $V_i$ and 
$\mathbf{\hat{w}}_j$ is the projected (or learned)
representation of word $w$ in space $V_j$. Squared error between 
$\mathbf{w}_j$ and
$\mathbf{\hat{w}}_j$ is the training loss to minimize. We use $\hat\mathbf{w}_j=f_{ij}(\mathbf{w}_i)$ to reformat
Equation \ref{equ:proj}.
Totally $c(c-1)/2$ projections $f_{ij}$ are trained, each on the
vocabulary intersection $V_{ij}$.

Let $w$ be a word that is unknown in $V_i$, but is known in
$V_1, V_2, \ldots, V_k$. To compute an embedding for $w$ in
$V_i$, we first compute the $k$ projections
$f_{1i}(\mathbf{w}_1)$, $f_{2i}(\mathbf{w}_2)$, $\ldots$,
$f_{ki}(\mathbf{w}_k)$ from the source spaces $V_1, V_2, \ldots, V_k$ to the target
space $V_i$. Then, the element-wise average of
$f_{1i}(\mathbf{w}_1)$, $f_{2i}(\mathbf{w}_2)$, $\ldots$,
$f_{ki}(\mathbf{w}_k)$ is treated as the representation of $w$ in $V_i$.
Our motivation is that -- assuming there is a
true representation of 
$w$ in $V_i$ (e.g., the one we would have obtained by
training embeddings on a much larger corpus) and assuming
the projections were learned well -- we would expect 
all the projected vectors
to be close to the true representation. Also,
each
source space contributes potentially complementary
information.  Hence averaging them is a balance of knowledge
from all source spaces.

%This
%projection idea between embedding spaces enables to learn
%representations of unknown words for an embedding set by
%projections from other sets. It holds that for each unknown
%word $w\in V_i^-$, we can find $k$ embedding sets $V^1, V^2,
%\ldots, V^k$ in which $w$ exists as a known word and its
%corresponding embeddings are $\mathbf{w}_1, \mathbf{w}_2,
%\ldots, \mathbf{w}_k$, respectively. We finally choose the
%\emph{element-wise average} of all $f_{1i}(\mathbf{w}_1)$,
%$f_{2i}(\mathbf{w}_2)$, $\ldots$, $f_{ki}(\mathbf{w}_k)$ as
%$\mathbf{w}_i$, the representation of unknown word $w$ in
%embedding set $V_i$. 

As discussed in Section \ref{sec:des}, we found that for the
binary sentiment classification dataset, many words were
unknown in at least one embedding version. But of these words, a
total of 5022 words did have coverage in another embedding
version and so will benefit from mutual-learning.  In the
experiments, we will show that this is a very effective
method to learn representations for unknown words that
increases system performance if learned representations are
used for initialization.

\textbf{Pretraining.}
Sentence classification systems are usually implemented as
supervised training regimes where training loss is between
true label distribution and predicted label distribution. In
this work, we use pretraining 
on the unlabeled data of each task and show that it can increase
the performance of classification systems.  

Figure \ref{fig:cnn} shows our pretraining setup. The
``sentence representation'' -- the output of ``Fully connected''
hidden layer -- is used to predict the component words
(``on'' in the figure) in
the sentence (instead of predicting the sentence label Y/N as in
supervised learning). Concretely, the sentence
representation is averaged with representations of some
surrounding words 
(``the'', ``cat'', ``sat'', ``the'', ``mat'', ``,'' in the figure)
to predict the middle word (``on'').

Given sentence representation $\mathbf{s}\in\mathbb{R}^d$
and initialized representations of $2t$ context words ($t$
left words and $t$ right words): $\mathbf{w}_{i-t}$,
$\ldots$, $\mathbf{w}_{i-1}$, $\mathbf{w}_{i+1}$, $\ldots$,
$\mathbf{w}_{i+t}$; $\mathbf{w}_i\in\mathbb{R}^d$, we
average the total $2t+1$ vectors element-wise, depicted as
``Average'' operation in Figure \ref{fig:cnn}. Then, this
resulting vector is treated as a predicted representation of
the middle word and is used to find the true middle word by
means of noise-contrastive estimation (NCE)
\cite{mnih2012fast}.  For each true example, 10 noise words
are sampled.

\iffalse NCE enables our model to learn to
discriminate between true middle words and noise words. NCE allows us to fit
unnormalized models making the training time effectively
independent of the vocabulary size.\fi

Note that in pretraining, there are three places where each
word needs initialization. (i)
Each word in the sentence is initialized in the ``Multichannel
input''
  layer to the whole network.
(ii) Each context word is initialized as input to the
average layer (``Average'' in the figure). (iii)
Each target word is initialized as the output of the ``NCE''
layer (``on'' in the figure).
 In this work, we use
multichannel initialization for case (i) and random
initialization for cases (ii) and (iii).  Only fine-tuned
multichannel representations (case (i)) are kept for
subsequent supervised training.

The rationale for this pretraining is similar to
auto-encoder: for an object composed of smaller-granular
elements, the representations of the whole object and its
components can learn each other. The CNN architecture learns
sentence features layer by layer, then those features are
justified by all constituent words.

During pretraining, all the model parameters, including
mutichannel input, convolution parameters and fully
connected layer, will be updated until they are mature to
extract the sentence features. Subsequently, the same sets
of parameters
will be fine-tuned for supervised
classification tasks.

In sum, this pretraining is designed to produce good initial
values for 
both model parameters and word
embeddings. It is especially helpful for pretraining the
embeddings of unknown words.

\begin{table*}[htbp]
\setlength{\tabcolsep}{2pt}
  \centering
  \begin{tabular}{c c r c c}\hline
    %\textbf{Systems}& \textbf{ROUGE-1} & \textbf{ROUGE-2} & \textbf{ROUGE-SU4}\\ \hline \hline
    Set & Training Data & Vocab Size & Dimensionality & Source\\ \hline \hline
    HLBL & Reuters English newswire & 246,122 & 50 & download\\
    Huang & Wikipedia (April 2010 snapshot) & 100,232 & 50 & download\\
    Glove & Twitter & 1,193,514 & 50 & download \\
    SENNA & Wikipedia & 130,000 & 50 & download\\
    Word2Vec & English Gigawords & 418,129 & 50 & trained from scratch\\\hline
  \end{tabular}
\caption{Description of five versions of word embedding.}\label{tab:sets}
\end{table*}

\section{Experiments}\label{sec:exp}
We test the network on four classification tasks. We begin by specifying aspects of the implementation and the training of the network. We then report the results of the experiments.

\subsection{Hyperparameters and Training}
In each of the experiments, the top  of the network
is a logistic regression
that predicts the probability distribution
over classes given the input sentence. The network is
trained to minimize cross-entropy of predicted and
true distributions; the objective includes an $L_2$
regularization term over the parameters. The set of
parameters comprises the word embeddings, all filter weights
and the weights in fully connected layers. A dropout
operation \cite{hinton2012improving} is put before the
logistic regression layer. The network is trained by
back-propagation in mini-batches and the gradient-based
optimization is performed using the AdaGrad update rule
\cite{duchi2011adaptive}

In all data sets, the initial learning rate is 0.01, dropout
probability is 0.8, $L_2$ weight is $5\cdot 10^{-3}$, batch size is
50. In each convolution layer, filter sizes are \{3, 5, 7,
9\} and each filter  has five kernels (independent of filter size).

\subsection{Datasets and Experimental Setup}

\textbf{Standard Sentiment Treebank}
\cite{socher2013recursive}. This small-scale dataset
includes two tasks predicting the sentiment of movie
reviews.  The output variable is binary in one experiment
and can have five possible outcomes in the other:
\{negative, somewhat negative, neutral, somewhat positive,
positive\}. In the \emph{binary} case, we use the given
split of 6920 training, 872 development and 1821 test
sentences. Likewise, in the \emph{fine-grained} case, we use
the standard 8544/1101/2210
split. \newcite{socher2013recursive} used the Stanford
Parser \cite{klein2003accurate} to parse each sentence into
subphrases. The subphrases were then labeled by human
annotators in the same way as the sentences were
labeled. Labeled phrases that occur as subparts of the
training sentences are treated as independent training
instances as in
\cite{le2014distributed,blunsom2014convolutional}.

\textbf{Sentiment140}\footnote{http://help.sentiment140.com/for-students}
\cite{go2009twitter}. This is a large-scale dataset of
tweets about sentiment classification, where a tweet is
automatically labeled as positive or negative depending on
the emoticon that occurs in it. The training set consists of
1.6 million tweets with emoticon-based labels and the test
set of about 400 hand-annotated tweets. We preprocess the
tweets minimally as follows. 1) The equivalence class symbol
``url'' (resp.\ ``username'') replaces all URLs (resp.\ 
all words that start
with the @ symbol, e.g., @thomasss).
2) A sequence of $k>2$ repetitions of a letter $c$ 
(e.g., ``cooooooool'')
is
replaced by two occurrences of $c$ (e.g., ``cool'').
3) All tokens
are lowercased.

% 1) Equivalence class tokens
%``username'' (resp.\ ``url'') replace all words that start
%with the @ symbol (e.g., @thomasss) (resp.\ replace all
%URLs). 2) A sequence of $k>2$ repetitions of a letter $c$ is
%replaced by two occurrences of $c$. For example,
%``huuuuungry'' is converted into ``huungry''. 3) All tokens
%are lowercased.

\textbf{Subj}. Subjectivity classification
dataset\footnote{http://www.cs.cornell.edu/people/pabo/movie-review-data/}
released by \cite{pang2004sentimental} has 5000 subjective
sentences and 5000 objective sentences. We report the result
of 10-fold cross validation as baseline systems did.

\begin{table}[htbp]
\setlength{\tabcolsep}{2pt}
  \centering
  \begin{tabular}{lrrrr}\hline
     & Binary  & Fine-grained  & Senti140  & Subj \\ \hline \hline
    HLBL & 5,232 & 5,562 & 344,632 & 8,621\\
    Huang & 4,273 & 4,523 & 327,067 & 6,382\\
    Glove & 3,299 & 3,485 & 257,376 & 5,237 \\
    SENNA & 4,136 & 4,371 & 323,501 & 6,162\\
    W2V & 2257 & 2,409 & 288,257 & 4,217\\\hline
    Voc size & 18,876 & 19,612 & 387,877 & 23,926\\
    Full hit & 12,030 & 12,357 & 30,010 & 13,742\\
    Partial hit & 5,022 & 5,312 & 121,383 & 6,580\\
    No hit & 1,824 & 1,943 & 236,484 & 3,604
  \end{tabular}
\caption{Statistics of five embedding versions for four
  tasks.  The first block with five rows provides the number
  of unknown words of each task when using corresponding version
  to initialize. 
Voc size:
vocabulary size.
Full
  hit: embedding in all 5 versions.
Partial hit: 
  embedding in 1--4 versions,
No hit: not present in any of the 5 versions.}
\label{tab:hits}
\end{table}
\subsubsection{Pretrained Word Vectors}

In this work, we use five embedding versions, as shown in \tabref{sets}, to initialize
words. Four of them are directly downloaded from the
Internet. (i) \textbf{HLBL.} Hierarchical log-bilinear model
presented by \newcite{mnih2009scalable} and released by
\newcite{turian2010word};\footnote{http://metaoptimize.com/projects/wordreprs/} size:
246,122 word embeddings; training corpus: RCV1 corpus,  one year of Reuters English newswire from August
1996 to August 1997. (ii) \textbf{Huang.}\footnote{
  http://ai.stanford.edu/~ehhuang/}
\newcite{huang2012improving} incorporated global context to
deal with challenges raised by words with multiple
meanings; size: 100,232 word embeddings; training corpus: 
April
2010 snapshot of 
Wikipedia. (iii)
\textbf{GloVe.}\footnote{http://nlp.stanford.edu/projects/glove/}
Size: 1,193,514 word embeddings; training corpus: a Twitter
corpus of
2B tweets with 27B tokens.
(iv) \textbf{SENNA.}\footnote{http://ml.nec-labs.com/senna/}
Size: 130,000 word embeddings; training corpus:  Wikipedia. Note
that we use their 50-dimensional
embeddings. (v) \textbf{Word2Vec.} It has no 50-dimensional
embeddings available online. We use released
code\footnote{http://code.google.com/p/word2vec/} to train
\emph{skip-gram} on English Gigaword Corpus
\cite{parker2009english} with setup: window size 5, negative
sampling, sampling rate $10^{-3}$, threads 12. It is worth emphasizing that above embeddings sets are derived on different corpora with different algorithms. This is the very property that we want to make use of to promote the system performance.

Table \ref{tab:hits} shows the number of unknown words in
each task when using corresponding embedding version to
initialize (rows ``HLBL'', ``Huang'', ``Glove'', ``SENNA'',
``W2V'')
and the number of words fully
initialized by five embedding versions (``Full hit'' row), the
number of words partially initialized (``Partial hit'' row) and
the number of words that cannot be initialized by any of
the embedding versions (``No hit'' row).

About 30\% of words in each task have partially initialized
embeddings and our mutual-learning is able to 
initialize the missing embeddings through
projections. Pretraining 
is expected to learn good
representations for all words, but pretraining is especially
important for 
words without initialization (``no hit''); a particularly
clear example for this is 
the Senti140 task: 
236,484 of 387,877 words or 61\% are in the ``no hit'' category.

\subsubsection{Results and Analysis}
\begin{table*}[htbp]
\setlength{\tabcolsep}{2pt}
  \centering
  \begin{tabular}{c r|l c c c c}\hline
    %\textbf{Systems}& \textbf{ROUGE-1} & \textbf{ROUGE-2} & \textbf{ROUGE-SU4}\\ \hline \hline
    &&Model & Binary & Fine-grained & Senti140 & Subj\\ \hline \hline
    \multirow{11}{*}{baselines}& 1&RAE \cite{socher2011semi} & 82.4 & 43.2 & -- &--\\
&2    &MV-RNN \cite{socher2012semantic} & 82.9 & 44.4 & -- &--\\
&3    &RNTN \cite{socher2013recursive} & 85.4 & 45.7 & --&--\\
&4    &DCNN \cite{blunsom2014convolutional} & 86.8 & 48.5  & \emph{87.4} &--\\
&5    &Paragraph-Vec \cite{le2014distributed} & 87.7 & \emph{48.7} &--&--\\
    
&6    &CNN-rand \cite{kimEMNLP2014} & 82.7& 45.0& --& 89.6\\
&7    &CNN-static \cite{kimEMNLP2014} & 86.8&45.5 & --& 93.0\\    
&8    &CNN-non-static \cite{kimEMNLP2014} & 87.2& 48.0 &-- & 93.4 \\    
&9    &CNN-multichannel \cite{kimEMNLP2014} & \emph{88.1} & 47.4 & -- & 93.2\\
    
&10    &NBSVM \cite{wang2012baselines} &--&--&--& 93.2\\
&11    &MNB \cite{wang2012baselines} &--&--&--& \emph{93.6}\\
&12    &G-Dropout \cite{wang2013fast} &--&--&--& 93.4 \\
&13    &F-Dropout \cite{wang2013fast} &--&--&--& \emph{93.6} \\

&14    &SVM \cite{go2009twitter} & -- & -- & 81.6 &--\\
&15    & BINB \cite{go2009twitter} & -- & -- & 82.7&--\\
&16    & MAX-TDNN \cite{blunsom2014convolutional} & -- & -- & 78.8&--\\
&17    &NBOW \cite{blunsom2014convolutional} & -- & -- & 80.9&--\\
&18    &MAXENT \cite{go2009twitter} & -- & -- & 83.0&--\\\hline
    \multirow{5}{*}{versions}&19&MVCNN (-HLBL) & 88.5& 48.7& 88.0& 93.6\\
&20    &MVCNN (-Huang) & 89.2 & 49.2& 88.1& 93.7\\ 
&21    &MVCNN (-Glove) & 88.3& 48.6& 87.4& 93.6\\
&22    &MVCNN (-SENNA) & 89.3& 49.1& 87.9& 93.4\\
&23    &MVCNN (-Word2Vec) & 88.4& 48.2& 87.6& 93.4\\\hline
    \multirow{4}{*}{filters}&24 & MVCNN (-3) & 89.1& 49.2& 88.0& 93.6\\
&25    & MVCNN (-5) & 88.7&49.0& 87.5& 93.4\\
&26    & MVCNN (-7) & 87.8&48.9&87.5 & 93.1\\
&27    & MVCNN (-9) & 88.6&49.2&87.8 & 93.3\\\hline
   \multirow{2}{*}{tricks}&28&MVCNN (-mutual-learning) & 88.2 & 49.2& 87.8& 93.5\\
& 29   &MVCNN (-pretraining) & 87.6 & 48.9& 87.6& 93.2\\\hline
   \multirow{4}{*}{layers} & 30&MVCNN (1) & 89.0& 49.3& 86.8& 93.8\\
&  31  & MVCNN (2) & \underline{89.4}& \underline{49.6}& 87.6& \underline{93.9}\\
&  32  & MVCNN (3) & 88.6& 48.6& \underline{88.2}& 93.1\\
&  33  & MVCNN (4) & 87.9& 48.2& 88.0& 92.4\\\hline
&  34  &MVCNN (overall) & $\textbf{89.4}$ &  $\textbf{49.6}$ & \textbf{88.2}&$\textbf{93.9}$ \\\hline
  \end{tabular}
\caption{Test set results of our CNN model against other
  methods. \textbf{RAE}: Recursive Autoencoders with
  pretrained word embeddings from Wikipedia
  \cite{socher2011semi}. \textbf{MV-RNN}: Matrix-Vector
  Recursive Neural Network with parse trees
  \cite{socher2012semantic}. \textbf{RNTN}: Recursive Neural
  Tensor Network with tensor-based feature function and
  parse trees \cite{socher2013recursive}. \textbf{DCNN,
    MAX-TDNN, NBOW}: Dynamic Convolution Neural Network
  with k-max pooling, Time-Delay Neural Networks with
  Max-pooling \cite{collobert2008unified}, Neural
  Bag-of-Words Models
  \cite{blunsom2014convolutional}. \textbf{Paragraph-Vec}:
  Logistic regression on top of paragraph vectors
  \cite{le2014distributed}. \textbf{SVM, BINB, MAXENT}:
  Support Vector Machines, Naive Bayes with unigram features
  and bigram features, Maximum Entropy
  \cite{go2009twitter}. \textbf{NBSVM, MNB}: Naive Bayes SVM
  and Multinomial Naive Bayes with uni-bigrams from
  \newcite{wang2012baselines}. \textbf{CNN-rand/static/multichannel/nonstatic}:
  CNN with word embeddings randomly initialized /
  initialized by pretrained vectors and kept static during
  training / initialized with two copies (each is a
  ``channel'') of pretrained embeddings / initialized with
  pretrained embeddings while fine-tuned during training
  \cite{kimEMNLP2014}. \textbf{G-Dropout, F-Dropout}:
  Gaussian Dropout and Fast Dropout from
  \newcite{wang2013fast}. Minus sign ``-'' in MVCNN (-Huang)
  etc. means ``Huang'' is not used. ``\textbf{versions / filters
    / tricks / layers}'' denote the MVCNN variants with
  different setups: discard certain embedding version / discard
  certain filter size / discard mutual-learning or
  pretraining / different numbers of convolution
  layer.}\label{tab:senti}
\end{table*}

Table \ref{tab:senti} compares results on test of MVCNN and its
variants with other baselines in the four sentence
classification tasks. 
Row
34, ``MVCNN (overall)'', shows performance of the best
configuration of MVCNN, optimized on dev.
This
version 
uses five versions of word embeddings, four 
filter sizes (3, 5, 7, 9), both mutual-learning and pretraining, three
convolution layers for Senti140 task and two convolution
layers for the other tasks.
Overall, our system gets the best results, beating all baselines.

The
table contains five  blocks from top to bottom. Each block
investigates one specific configurational aspect of the system. All
results in the five blocks are with respect to row 34,
``MVCNN (overall)''; e.g., row 19 shows what happens when
HLBL is removed from row 34, row 28 shows what happens when
mutual learning is removed from row 34 etc.

The block ``baselines'' (1--18) lists some systems representative of previous
work on the corresponding datasets, including the
state-of-the-art systems (marked as italic). The block ``versions''
(19--23) shows
the results of our system when one of the embedding versions was \emph{not}
used during training. We want to explore to what extend
different embedding versions contribute to performance. The
block ``filters'' (24--27) gives the results when
individual filter width is discarded. It also tells us how
much a filter with specific size influences. The block ``tricks'' (28--29) shows the
system performance when no mutual-learning or no
pretraining is used. The block ``layers'' (30--33) demonstrates how the system performs
when it has different numbers of convolution layers. 

From the ``layers'' block, we can see that our system
performs best with two layers of convolution in Standard
Sentiment Treebank and Subjectivity Classification tasks
(row 31), but
with three layers of convolution in Sentiment140 (row 32). This
is probably due to Sentiment140 being a much larger
dataset; in such a case deeper neural networks are
beneficial. 

The block ``tricks'' demonstrates the effect of
mutual-learning and pretraining. Apparently,
pretraining has a bigger impact on performance than
mutual-learning. We speculate that it is because
pretraining can influence more words and all learned word
embeddings are tuned on the dataset after pretraining. 

The block ``filters'' indicates
the contribution of each filter size. The system benefits from filters of each size. Sizes 5 and 7 are most
important for high performance,
especially 7 (rows 25 and 26).

In the block
``versions'', we see that each
embedding version is crucial for good performance: performance drops in every single case. 
Though it is
not easy to compare fairly different embedding versions in NLP tasks,
especially when those embeddings were trained on different
corpora of different sizes using different algorithms, our
results are potentially instructive for researchers making
decision on which embeddings to use for their own tasks.

\section{Conclusion}\label{sec:conc}
This work presented  \emph{MVCNN}, a novel 
CNN architecture for sentence classification.  
It combines 
multichannel initialization
--
diverse versions of pretrained
word embeddings are used --
and variable-size filters --
features of multigranular
phrases are extracted with variable-size convolution
filters.  
We demonstrated that 
multichannel initialization
and variable-size filters enhance system performance
on sentiment
classification and subjectivity classification tasks.

\section{Future Work}
As pointed out by the reviewers the success of the
multichannel approach is likely due to a combination of
several quite different effects.

First, there is the effect of the embedding learning
algorithm. These algorithms differ in 
many aspects, including in
sensitivity to word
order (e.g., SENNA: yes, word2vec: no), in objective
function and in their treatment of ambiguity (explicitly
modeled only by 
\newcite{huang2012improving}.

Second, there is the effect of the corpus. We would expect the
size and genre of the corpus to have a big effect even
though we did not analyze this effect in this paper.

Third, complementarity of word embeddings is likely to be
more useful for some tasks than for others. Sentiment is a
good application for complementary word embeddings because
solving this task requires drawing on heterogeneous sources
of information, including syntax, semantics and genre as
well as the core polarity of a word. Other tasks like part
of speech (POS) tagging may benefit less from heterogeneity since
the benefit of embeddings in POS often comes down to making
a correct choice between two alternatives -- a single
embedding version may be sufficient for this.

We plan to pursue these questions in future work.

\section*{Acknowledgments}
Thanks to CIS members and anonymous reviewers for
constructive comments. This work was supported by Baidu
(through a Baidu scholarship awarded to Wenpeng Yin) and by
Deutsche Forschungsgemeinschaft (grant DFG SCHU 2246/8-2,
SPP 1335).

\bibliographystyle{acl}
\bibliography{acl2015}

\begin{thebibliography}{}

\bibitem[\protect\citename{Bengio \bgroup et al.\egroup
  }2003]{bengio2003neural}
Yoshua Bengio, R{\'e}jean Ducharme, Pascal Vincent, and Christian Janvin.
\newblock 2003.
\newblock A neural probabilistic language model.
\newblock {\em The Journal of Machine Learning Research}, 3:1137--1155.

\bibitem[\protect\citename{Blacoe and Lapata}2012]{blacoe2012comparison}
William Blacoe and Mirella Lapata.
\newblock 2012.
\newblock A comparison of vector-based representations for semantic
  composition.
\newblock In {\em Proceedings of the 2012 Joint Conference on Empirical Methods
  in Natural Language Processing and Computational Natural Language Learning},
  pages 546--556. Association for Computational Linguistics.

\bibitem[\protect\citename{Chen \bgroup et al.\egroup
  }2013]{chen2013expressive}
Yanqing Chen, Bryan Perozzi, Rami Al-Rfou, and Steven Skiena.
\newblock 2013.
\newblock The expressive power of word embeddings.
\newblock In {\em ICML Workshop on Deep Learning for Audio, Speech, and
  Language Processing}.

\bibitem[\protect\citename{Collobert and Weston}2008]{collobert2008unified}
Ronan Collobert and Jason Weston.
\newblock 2008.
\newblock A unified architecture for natural language processing: Deep neural
  networks with multitask learning.
\newblock In {\em Proceedings of the 25th international conference on Machine
  learning}, pages 160--167. ACM.

\bibitem[\protect\citename{Collobert \bgroup et al.\egroup
  }2011]{collobert2011natural}
Ronan Collobert, Jason Weston, L{\'e}on Bottou, Michael Karlen, Koray
  Kavukcuoglu, and Pavel Kuksa.
\newblock 2011.
\newblock Natural language processing (almost) from scratch.
\newblock {\em The Journal of Machine Learning Research}, 12:2493--2537.

\bibitem[\protect\citename{Denil \bgroup et al.\egroup
  }2014]{denil2014modelling}
Misha Denil, Alban Demiraj, Nal Kalchbrenner, Phil Blunsom, and Nando
  de~Freitas.
\newblock 2014.
\newblock Modelling, visualising and summarising documents with a single
  convolutional neural network.
\newblock {\em arXiv preprint arXiv:1406.3830}.

\bibitem[\protect\citename{Dos~Santos and Gatti}2014]{dos2014deep}
C{\i}cero~Nogueira Dos~Santos and Ma{\i}ra Gatti.
\newblock 2014.
\newblock Deep convolutional neural networks for sentiment analysis of short
  texts.
\newblock In {\em Proceedings of the 25th International Conference on
  Computational Linguistics}.

\bibitem[\protect\citename{Duchi \bgroup et al.\egroup
  }2011]{duchi2011adaptive}
John Duchi, Elad Hazan, and Yoram Singer.
\newblock 2011.
\newblock Adaptive subgradient methods for online learning and stochastic
  optimization.
\newblock {\em The Journal of Machine Learning Research}, 12:2121--2159.

\bibitem[\protect\citename{Go \bgroup et al.\egroup }2009]{go2009twitter}
Alec Go, Richa Bhayani, and Lei Huang.
\newblock 2009.
\newblock Twitter sentiment classification using distant supervision.
\newblock {\em CS224N Project Report, Stanford}, pages 1--12.

\bibitem[\protect\citename{Graves \bgroup et al.\egroup
  }2013]{graves2013speech}
Alex Graves, A-R Mohamed, and Geoffrey Hinton.
\newblock 2013.
\newblock Speech recognition with deep recurrent neural networks.
\newblock In {\em Acoustics, Speech and Signal Processing, 2013 IEEE
  International Conference on}, pages 6645--6649. IEEE.

\bibitem[\protect\citename{Hill \bgroup et al.\egroup }2014]{hill2014not}
Felix Hill, KyungHyun Cho, Sebastien Jean, Coline Devin, and Yoshua Bengio.
\newblock 2014.
\newblock Not all neural embeddings are born equal.
\newblock In {\em NIPS Workshop on Learning Semantics}.

\bibitem[\protect\citename{Hinton \bgroup et al.\egroup
  }2012]{hinton2012improving}
Geoffrey~E Hinton, Nitish Srivastava, Alex Krizhevsky, Ilya Sutskever, and
  Ruslan~R Salakhutdinov.
\newblock 2012.
\newblock Improving neural networks by preventing co-adaptation of feature
  detectors.
\newblock {\em arXiv preprint arXiv:1207.0580}.

\bibitem[\protect\citename{Huang \bgroup et al.\egroup
  }2012]{huang2012improving}
Eric~H Huang, Richard Socher, Christopher~D Manning, and Andrew~Y Ng.
\newblock 2012.
\newblock Improving word representations via global context and multiple word
  prototypes.
\newblock In {\em Proceedings of the 50th Annual Meeting of the Association for
  Computational Linguistics: Long Papers-Volume 1}, pages 873--882. Association
  for Computational Linguistics.

\bibitem[\protect\citename{Kalchbrenner \bgroup et al.\egroup
  }2014]{blunsom2014convolutional}
Nal Kalchbrenner, Edward Grefenstette, and Phil Blunsom.
\newblock 2014.
\newblock A convolutional neural network for modelling sentences.
\newblock In {\em Proceedings of the 52nd Annual Meeting of the Association for
  Computational Linguistics}. Association for Computational Linguistics.

\bibitem[\protect\citename{Kim}2014]{kimEMNLP2014}
Yoon Kim.
\newblock 2014.
\newblock Convolutional neural networks for sentence classification.
\newblock In {\em Proceedings of the 2014 Conference on Empirical Methods in
  Natural Language Processing}, October.

\bibitem[\protect\citename{Klein and Manning}2003]{klein2003accurate}
Dan Klein and Christopher~D Manning.
\newblock 2003.
\newblock Accurate unlexicalized parsing.
\newblock In {\em Proceedings of the 41st Annual Meeting on Association for
  Computational Linguistics-Volume 1}, pages 423--430. Association for
  Computational Linguistics.

\bibitem[\protect\citename{Krizhevsky \bgroup et al.\egroup
  }2012]{krizhevsky2012imagenet}
Alex Krizhevsky, Ilya Sutskever, and Geoffrey~E Hinton.
\newblock 2012.
\newblock Imagenet classification with deep convolutional neural networks.
\newblock In {\em Advances in neural information processing systems}, pages
  1097--1105.

\bibitem[\protect\citename{Le and Mikolov}2014]{le2014distributed}
Quoc~V Le and Tomas Mikolov.
\newblock 2014.
\newblock Distributed representations of sentences and documents.
\newblock In {\em Proceedings of the 31st international conference on Machine
  learning}.

\bibitem[\protect\citename{Luo \bgroup et al.\egroup }2014]{luo2014pre}
Yong Luo, Jian Tang, Jun Yan, Chao Xu, and Zheng Chen.
\newblock 2014.
\newblock Pre-trained multi-view word embedding using two-side neural network.
\newblock In {\em Twenty-Eighth AAAI Conference on Artificial Intelligence}.

\bibitem[\protect\citename{Mikolov \bgroup et al.\egroup
  }2010]{mikolov2010recurrent}
Tomas Mikolov, Martin Karafi{\'a}t, Lukas Burget, Jan Cernock{\`y}, and Sanjeev
  Khudanpur.
\newblock 2010.
\newblock Recurrent neural network based language model.
\newblock In {\em INTERSPEECH 2010, 11th Annual Conference of the International
  Speech Communication Association, Makuhari, Chiba, Japan, September 26-30,
  2010}, pages 1045--1048.

\bibitem[\protect\citename{Mikolov \bgroup et al.\egroup
  }2013a]{mikolov2013efficient}
Tomas Mikolov, Kai Chen, Greg Corrado, and Jeffrey Dean.
\newblock 2013a.
\newblock Efficient estimation of word representations in vector space.
\newblock In {\em Proceedings of Workshop at ICLR}.

\bibitem[\protect\citename{Mikolov \bgroup et al.\egroup
  }2013b]{mikolov2013distributed}
Tomas Mikolov, Ilya Sutskever, Kai Chen, Greg~S Corrado, and Jeff Dean.
\newblock 2013b.
\newblock Distributed representations of words and phrases and their
  compositionality.
\newblock In {\em Advances in Neural Information Processing Systems}, pages
  3111--3119.

\bibitem[\protect\citename{Mikolov}2012]{mikolov2012statistical}
Tomas Mikolov.
\newblock 2012.
\newblock Statistical language models based on neural networks.
\newblock {\em Presentation at Google, Mountain View, 2nd April}.

\bibitem[\protect\citename{Mnih and Hinton}2009]{mnih2009scalable}
Andriy Mnih and Geoffrey~E Hinton.
\newblock 2009.
\newblock A scalable hierarchical distributed language model.
\newblock In {\em Advances in neural information processing systems}, pages
  1081--1088.

\bibitem[\protect\citename{Mnih and Teh}2012]{mnih2012fast}
Andriy Mnih and Yee~Whye Teh.
\newblock 2012.
\newblock A fast and simple algorithm for training neural probabilistic
  language models.
\newblock In {\em Proceedings of the 29th International Conference on Machine
  Learning}, pages 1751--1758.

\bibitem[\protect\citename{Pang and Lee}2004]{pang2004sentimental}
Bo~Pang and Lillian Lee.
\newblock 2004.
\newblock A sentimental education: Sentiment analysis using subjectivity
  summarization based on minimum cuts.
\newblock In {\em Proceedings of the 42nd annual meeting on Association for
  Computational Linguistics}, page 271. Association for Computational
  Linguistics.

\bibitem[\protect\citename{Parker \bgroup et al.\egroup
  }2009]{parker2009english}
Robert Parker, Linguistic~Data Consortium, et~al.
\newblock 2009.
\newblock {\em English gigaword fourth edition}.
\newblock Linguistic Data Consortium.

\bibitem[\protect\citename{Pennington \bgroup et al.\egroup
  }2014]{pennington2014glove}
Jeffrey Pennington, Richard Socher, and Christopher~D Manning.
\newblock 2014.
\newblock Glove: Global vectors for word representation.
\newblock {\em Proceedings of the Empiricial Methods in Natural Language
  Processing}, 12.

\bibitem[\protect\citename{Socher \bgroup et al.\egroup
  }2011a]{socher2011dynamic}
Richard Socher, Eric~H Huang, Jeffrey Pennin, Christopher~D Manning, and
  Andrew~Y Ng.
\newblock 2011a.
\newblock Dynamic pooling and unfolding recursive autoencoders for paraphrase
  detection.
\newblock In {\em Advances in Neural Information Processing Systems}, pages
  801--809.

\bibitem[\protect\citename{Socher \bgroup et al.\egroup }2011b]{socher2011semi}
Richard Socher, Jeffrey Pennington, Eric~H Huang, Andrew~Y Ng, and
  Christopher~D Manning.
\newblock 2011b.
\newblock Semi-supervised recursive autoencoders for predicting sentiment
  distributions.
\newblock In {\em Proceedings of the Conference on Empirical Methods in Natural
  Language Processing}, pages 151--161. Association for Computational
  Linguistics.

\bibitem[\protect\citename{Socher \bgroup et al.\egroup
  }2012]{socher2012semantic}
Richard Socher, Brody Huval, Christopher~D Manning, and Andrew~Y Ng.
\newblock 2012.
\newblock Semantic compositionality through recursive matrix-vector spaces.
\newblock In {\em Proceedings of the 2012 Joint Conference on Empirical Methods
  in Natural Language Processing and Computational Natural Language Learning},
  pages 1201--1211. Association for Computational Linguistics.

\bibitem[\protect\citename{Socher \bgroup et al.\egroup
  }2013]{socher2013recursive}
Richard Socher, Alex Perelygin, Jean~Y Wu, Jason Chuang, Christopher~D Manning,
  Andrew~Y Ng, and Christopher Potts.
\newblock 2013.
\newblock Recursive deep models for semantic compositionality over a sentiment
  treebank.
\newblock In {\em Proceedings of the conference on empirical methods in natural
  language processing}, volume 1631, page 1642. Citeseer.

\bibitem[\protect\citename{Turian \bgroup et al.\egroup }2010]{turian2010word}
Joseph Turian, Lev Ratinov, and Yoshua Bengio.
\newblock 2010.
\newblock Word representations: a simple and general method for semi-supervised
  learning.
\newblock In {\em Proceedings of the 48th annual meeting of the association for
  computational linguistics}, pages 384--394. Association for Computational
  Linguistics.

\bibitem[\protect\citename{Wang and Manning}2012]{wang2012baselines}
Sida Wang and Christopher~D Manning.
\newblock 2012.
\newblock Baselines and bigrams: Simple, good sentiment and topic
  classification.
\newblock In {\em Proceedings of the 50th Annual Meeting of the Association for
  Computational Linguistics: Short Papers-Volume 2}, pages 90--94. Association
  for Computational Linguistics.

\bibitem[\protect\citename{Wang and Manning}2013]{wang2013fast}
Sida Wang and Christopher Manning.
\newblock 2013.
\newblock Fast dropout training.
\newblock In {\em Proceedings of the 30th International Conference on Machine
  Learning}, pages 118--126.

\bibitem[\protect\citename{Yin and Sch{\"u}tze}2014]{yin2014exploration}
Wenpeng Yin and Hinrich Sch{\"u}tze.
\newblock 2014.
\newblock An exploration of embeddings for generalized phrases.
\newblock {\em Proceedings of the 52nd annual meeting of the association for
  computational linguistics, student research workshop}, pages 41--47.

\bibitem[\protect\citename{Yu \bgroup et al.\egroup }2014]{yu2014deep}
Lei Yu, Karl~Moritz Hermann, Phil Blunsom, and Stephen Pulman.
\newblock 2014.
\newblock Deep learning for answer sentence selection.
\newblock {\em NIPS deep learning workshop}.

\end{thebibliography}
\end{document}